\documentclass[final]{cvpr}

\usepackage{times}
\usepackage{epsfig}
\usepackage{graphicx}
\usepackage{amsmath}
\usepackage{amssymb}
\usepackage{color}
\usepackage{cuted}
\usepackage{capt-of}
\usepackage{booktabs}  
\usepackage{multirow}

\usepackage{pifont}
\makeatletter
\renewcommand*\env@matrix[1][*\c@MaxMatrixCols c]{%
   \hskip -\arraycolsep
   \let\@ifnextchar\new@ifnextchar
   \array{#1}}
\makeatother

\usepackage[pagebackref=true,breaklinks=true,letterpaper=true,colorlinks,bookmarks=false]{hyperref}

\def\cvprPaperID{5496} %
\def\confYear{CVPR 2021}

\begin{document}

\title{LASR: Learning Articulated Shape Reconstruction from a Monocular Video\vspace{-1ex}}

\author{Gengshan Yang$^{1\thanks{Work done in an internship at Google. }}$,\, Deqing Sun$^{2}$, \, Varun Jampani$^{2}$, \, Daniel Vlasic$^{2}$, \, Forrester Cole$^{2}$, \, Huiwen Chang$^{2}$,\\ Deva Ramanan$^{1}$, \, William T. Freeman$^{2}$, and Ce Liu$^{2}$\\
\vspace{1ex}
{$^1$Carnegie Mellon University,\quad $^2$Google Research}
\vspace{-4ex}
}

\maketitle

\begin{strip}\centering
\vspace{-30pt}
\includegraphics[width=\linewidth]{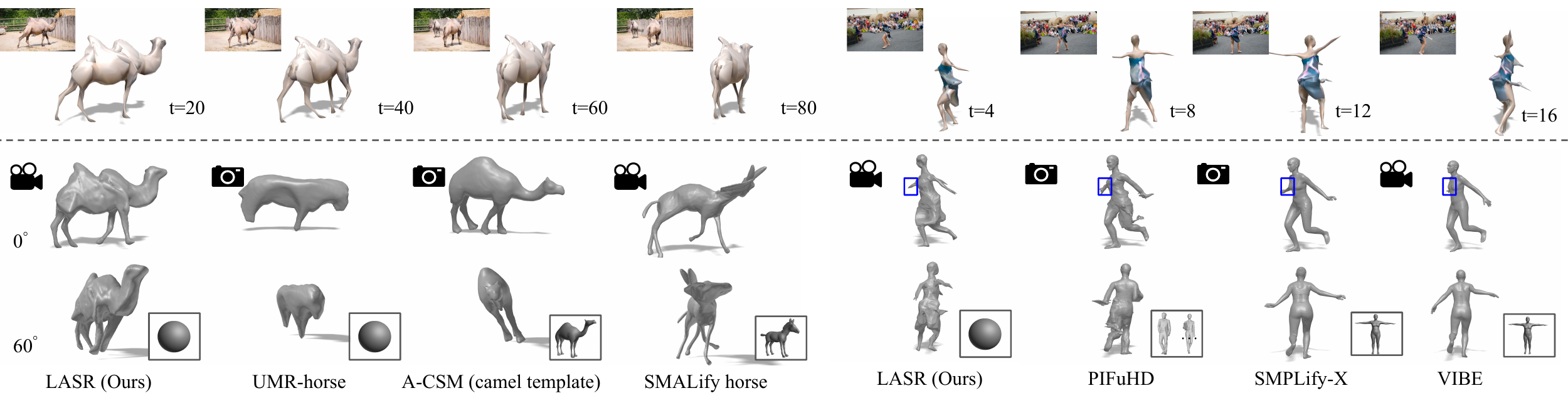}
\captionof{figure}{{\bf Top}: Sample input video frames and articulated shapes recovered by our method (LASR). {\bf Bottom}: Comparison with existing methods, where the input to each method (either video or image) is denoted at the top left, and the shape template being used is denoted at the bottom right of each result. Many existing approaches on nonrigid shape reconstruction heavily rely on category-specific 3D shape templates, such as SMPL for human~\cite{SMPL:2015, SMPL-X:2019} and SMAL for quadrupeds~\cite{biggs2018creatures, Zuffi:CVPR:2017}. In contrast, LASR jointly recovers the object shape, articulation, and camera parameters from a monocular video without using category-specific shape templates. By relying on generic shape and motion priors, LASR applies to a wider range of nonrigid shapes and yields high-fidelity 3D reconstructions: It recovers both humps of the camel, which are missing from other methods. It also recovers the silk ribbon of the dancer (as denoted by the blue box), which confuses SMPLify-X and VIBE as the right arm. Please refer to video results on our project page.
\label{fig:teaser}}
\end{strip}

\begin{abstract}
Remarkable progress has been made in 3D reconstruction of rigid structures from a video or a collection of images. However, it is still challenging to reconstruct nonrigid structures from RGB inputs, due to its under-constrained nature. While template-based approaches, such as parametric shape models, have achieved great success in modeling the ``closed world" of known object categories, they cannot well handle the ``open-world" of novel object categories or outlier shapes. In this work, we introduce a template-free approach to learn 3D shapes from a single video. It adopts an analysis-by-synthesis strategy that forward-renders object silhouette, optical flow, and pixel values to compare with video observations, which generates gradients to adjust the camera, shape and motion parameters. Without using a category-specific shape template, our method faithfully reconstructs nonrigid 3D structures from videos of human, animals, and objects of unknown classes. Our code is available at \href{lasr-google.github.io}{\url{lasr-google.github.io}}.
\end{abstract}

\section{Introduction}
Perceiving and modeling the geometry and dynamics of 3D entities is an open research problem in computer vision and has numerous applications. One fundamental challenge is the under-constrained nature of the problem: from limited 2D image measurements, there exist multiple interpretations of the geometry and motion of the 3D world. 

A recent and promising trend for addressing this challenge is exploiting data priors, which have proven
quite successful for high-level vision tasks, such as image
classification and object detection~\cite{deng2009imagenet,lin2014microsoft}. However, in contrast to high-level vision tasks, it is often costly to obtain 3D annotations for real-world entities. For example, SMPL~\cite{SMPL:2015} is learned from thousands of registered 3D scans of human. SMAL~\cite{Zuffi:CVPR:2017} is learned from scans of animal toys and a manually rigged mesh model. It involves nontrivial efforts to collect such data for an arbitrary object category. Therefore, existing methods often fail to capture objects of novel or unknown classes, and hallucinate an average 3D structure based on the category shape prior, as shown in Fig.~\ref{fig:teaser}.

Interestingly, remarkable progress has been made in the field of SLAM and structure-from-motion without relying on strong shape priors by taking advantage of {\em multiview} data recordings. However, such results are limited to static scenes. We explore an intermediate regime between these two extremes: {\em Can one reconstruct an articulated shape from video data without relying on template priors?} 

\noindent\textbf{Why videos?}
To reconstruct 3D object shape from images, prior work learns category-specific shape models either from 3D data~\cite{gkioxari2019mesh,saito2020pifuhd} or from 2D supervision, such as object silhouette and keypoints in a large image collection~\cite{cashman2012shape,ucmrGoel20, cmrKanazawa18,li2020self}. However, 3D data are generally difficult to acquire at a large scale due to sensor design. Although it is easier to collect images of the same category, enforcing multiview constraints is often challenging, due to ambiguities of associating 2D observations across instances and under different viewpoints~\cite{choy_nips16, NEURIPS2018_8f7d807e}. Video serves as an alternative to depth scans and image collections -- videos are easier to acquire, and provide well-defined multiview constraints on the 3D shape of the same instance.

\noindent\textbf{Why optical flow?}
To solve the inverse problem, prior work discussed various forms of 2D constraints or supervision, such as object silhouette, texture, 2D keypoints, and semantic parts~\cite{baker2005shape,ucmrGoel20,cmrKanazawa18,li2020self}. Why should motion be treated as a first-class citizen? Besides that optical flow naturally encodes correspondences, it provides more fine-grained information than keypoints as well as semantic parts. Different from long-range point tracks, which is the classic input for NRSfM~\cite{sand2008particle}, optical flow can be obtained more reliably~\cite{teed2020raft,yang2019volumetric} over two consecutive frames. 

\noindent\textbf{Why not nonrigid SfM?}
NRSfM deals with a problem similar to ours: given a set of 2D point trajectories depicting a deformable object in a collection of images, the goal is to recover the 3D object shape and pose (i.e., relative camera position) in each view. Usually, trajectories of 2D points are factorized into low-rank shape and motion matrices~\cite{bregler2000recovering,gotardo2011non,kong2019deep} without using 3D shape templates. Although NRSfM is able to deal with generic shapes, it requires reliable long-term point tracks or keypoint annotations, which are challenging to acquire densely in practice~\cite{sand2008particle,sidhu2020neural,sundaram2010dense}. 

\begin{figure*}
    \centering
    \includegraphics[width=0.85\linewidth]{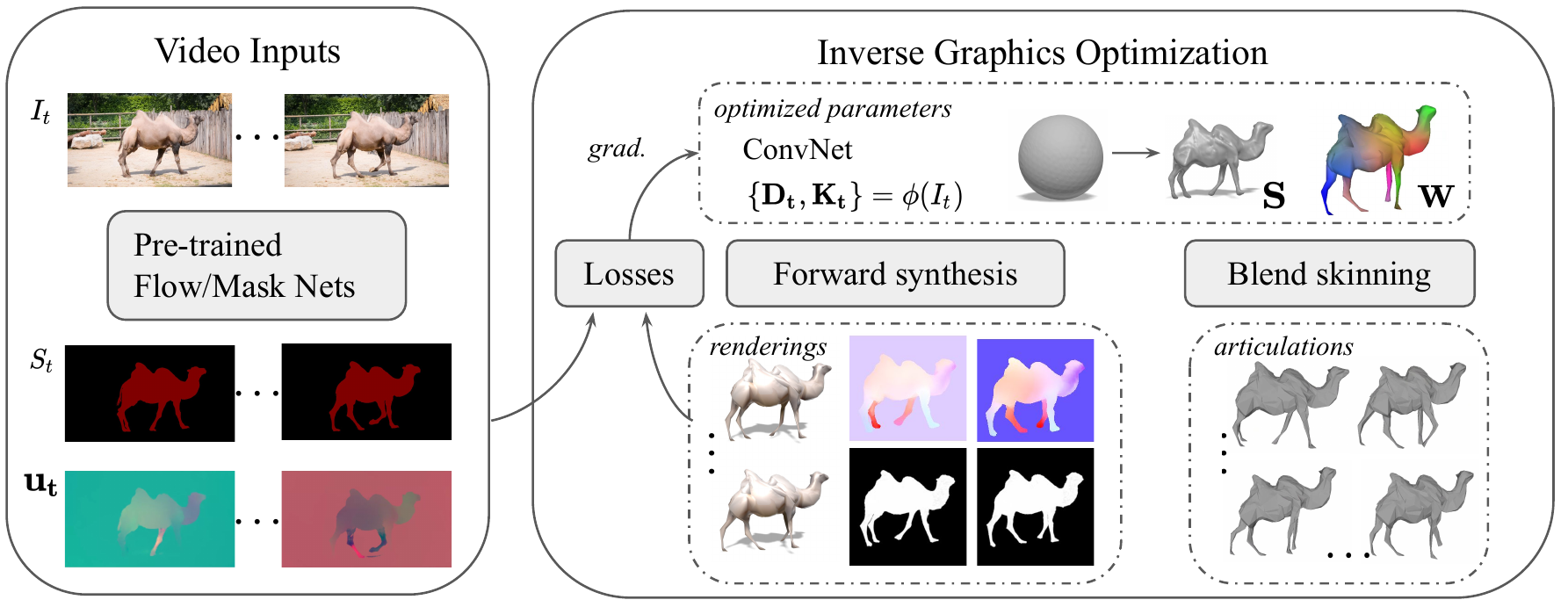}
    \caption{Method overview. Given a monocular video of an object, we jointly recover the object's rest shape ${\bf S}$, skinning weights ${\bf W}$, articulation ${\bf D_t}$, and camera parameters ${\bf K_t}$ by solving an inverse graphics problem through gradient-based optimization. The object rest shape is represented by a mesh (initialized from a sphere) and articulated at each frame under linear blend skinning (Sec.~\ref{sec:deformation}). The time-varying parameters, including focal length, object root transformation and articulation, are parameterized by a neural basis, i.e., convolutional network given image inputs  (Sec.~\ref{sec:details}).  At each iteration, we randomly sample pairs of consecutive frames and forward-render object silhouette, texture, and two-frame optical flow using a differentiable renderer (Sec.~\ref{sec:model}).  The renderings are compared against raw pixels, segmentation and optical flow estimated by off-the-shelf networks to generate gradients signals and update the model (Sec.~\ref{sec:learning}).}
    \label{fig:overview}
\end{figure*}

\noindent\textbf{Proposed approach:}
Instead of inferring 3D shapes from category-specific image collections or point trajectories, we build an articulated shape model from a monocular video of an object. Recent progress in differentiable rendering allows one to recast the problem as an analysis-by-synthesis task: {\em we solve the inverse graphics problem of recovering the 3D object shape (including spacetime deformations) and camera trajectories (including intrinsics) to fit video observations, such as object silhouette, raw pixels, and optical flow}. An overview of the pipeline is shown in Fig.~\ref{fig:overview}.

\noindent\textbf{Contributions:} We propose a method for articulated shape reconstruction from a monocular video that does not require a prior template or category information. It takes advantage of dense two-frame optical flow 
to overcome the inherent ambiguity in the nonrigid structure and motion estimation problem. It automatically recovers a nonrigid shape under the constraints of rigid bones under linear-blend skinning. It combines coarse-to-fine re-meshing with soft-symmetric constraints to recover high-quality meshes. Our method demonstrates state-of-the-art reconstruction performance in the BADJA animal video dataset~\cite{biggs2018creatures}, 
strong performance against model-based methods on humans, and higher accuracy on two animated animals than A-CSM~\cite{kulkarni2020articulation} and SMALify~\cite{biggs2018creatures} 
that use shape templates.

\begin{table}[!t]
    \caption{Related work in nonrigid shape reconstruction. $^{(1)}$Model-based optimization and regression methods. $^{(2)}$Category-specific mesh reconstruction. $^{(3)}$Template-free approaches. S: single view. V: video or multi-view data. I: images. J2: 2D joints. J3: 3D joints. M: 2D masks. V3: 3D scans. C: camera matrices. F: optical flow. MF: multi-frame optical flow. quad: quadruped animals. $^\dagger$:Only representative categories are listed.}
    \footnotesize
    \centering
    \begin{tabular}{rrcccccccc}
	\toprule
 &Method  &  category & template & test-input & train\\
 \midrule
 \multirow{6}{*}{\shortstack{(1)}}
 &SMPLify~\cite{Bogo:ECCV:2016}   & human  & SMPL  & S:J2,M & None\\
  &VIBE~\cite{kocabas2019vibe}   & human                         & SMPL  & V:I   & J2,J3\\
 &SMALify~\cite{biggs2018creatures} & quadx5                   & SMAL  & V:J2,M & None\\
 &SMALR~\cite{Zuffi:CVPR:2018}  & quadx12                      & SMAL  & S:J2,M & None\\
 &SMALST~\cite{zuffi2019three} & zebra            & SMAL  & S:I   & J2,V3\\
 &WLDO~\cite{biggs2020wldo}   & dog                          & SMAL  & S:I    & J2,M\\
 \midrule
  \multirow{6}{*}{\shortstack{(2)}}
 &CMR~\cite{cmrKanazawa18}   & bird$^\dagger$                & SfM-hull  & S:I & J2,M,C\\
 &UCMR~\cite{ucmrGoel20}   & bird$^\dagger$           & cate-mesh & S:I & M\\
 &UMR~\cite{li2020self}    & bird$^\dagger$         & None          & S:I & M\\
 &IMR~\cite{tulsiani2020imr}    & animals           & cate-mesh & S:I & M\\
 &A-CSM~\cite{kulkarni2020articulation}  & animals         & cate-mesh & S:I & M\\
 &VMR~\cite{vmr2020}    & animals  & cate-mesh & V:M & None \\
 \midrule
  \multirow{6}{*}{\shortstack{(3)}}
 &PIFuHD~\cite{saito2020pifuhd} & human             & None  & S:I & V3 \\
 &NRSfM~\cite{agudo2018image,dai2014simple}  & any               & None  & V:J2 & None\\
 &N-NRSfM~\cite{sidhu2020neural}  & any               & None  & V:MF,M & None\\
 &WSD~\cite{cashman2012shape}   & dolphin$^\dagger$   & cate-mesh    & V:J2,M & None\\
  &A3DC~\cite{Reinert:2016:10.20380/GI2016.17}   & any   & None         & V:stroke        & None\\ 
 &LASR (Ours)   & any   & None    & V:F,M    & None \\
 \bottomrule
\label{tab:taxonomy}
\end{tabular}
\vspace{-20pt}
\end{table}

\section{Related Work}

Below and in Tab.~\ref{tab:taxonomy}, we discuss related work of nonrigid shape recovery according to priors being used (shape template, category prior, or generic shape and motion priors).

\noindent\textbf{Model-based reconstruction:}
Model-based reconstruction leverages a parametric shape model to solve the under-constrained 3D shape and pose estimation problem. A large body of work in 3D human and animal reconstruction uses such parametric shape models~\cite{SMPL:2015,SMPL-X:2019,xiang2019monocular,Zuffi:CVPR:2018,Zuffi:CVPR:2017}, which are learned from 3D scans of human or animal toys~\cite{SMPL:2015,Zuffi:CVPR:2017}, and allow one to recover the 3D shape given very few annotations at test time (2D keypoints and silhouettes). Recently, model-based regression methods are developed to predict model-specific shape and pose parameters from a single image or video~\cite{badger2020,biggs2020wldo,kocabas2019vibe,zuffi2019three}, usually trained with ground-truth 3D data generated by such parametric shape models. Although model-based methods achieve great success in reconstructing ``closed-world'' objects of known category and rich 3D data, it is challenging to apply to unknown object categories, or categories with limited 3D data. 

\noindent\textbf{Category mesh reconstruction:}
A recent trend is to reduce supervision from 3D or multi-view capturing to 2D annotations, such as keypoints and object silhouettes~\cite{ucmrGoel20,cmrKanazawa18,li2020self}. Such methods often take advantage of category priors, including a collection of images from the same category, and category-specific shape templates~\cite{kulkarni2020articulation,tulsiani2020imr}. Recent progress makes single-view reconstruction of birds and other common categories possible without 3D annotation. However, the single view reconstruction is usually coarse and lacks instance-specific details. Recent work adapts category-specific models to a test-time video~\cite{vmr2020}, but still does not handle objects of unknown classes.

\noindent\textbf{Template-free reconstruction:}
Among the template-free methods, PIFu~\cite{saito2019pifu,saito2020pifuhd} learns to predict an implicit shape representation for clothed human reconstruction, but requires ground-truth 3D shapes to train. A3DC~\cite{Reinert:2016:10.20380/GI2016.17} reconstructs articulated animals from videos, but requires involved user stroke interactions. Without requiring 3D data or user interactions, NRSfM factorizes a set of 2D keypoints or point trajectories into the 3D object shape and pose in each view assuming ``low-rank'' shape or deformation~\cite{agudo2018image, dai2014simple, gotardo2011non}. Recently, deep networks have been applied to learn such factorization of specific categories from 2D annotations~\cite{kong2019deep, novotny2019c3dpo, wang2020deep}. Close to our approach, Neural Dense NRSfM (N-NRSfM)~\cite{sidhu2020neural} learns a video-specific shape and deformation model from dense 2D point tracks. However, such methods are limited by the accuracy of 2D trajectory inputs, which is challenging to estimate in real-world sequences when large motion occurs~\cite{sand2008particle,sidhu2020neural,sundaram2010dense}.

\section{Approach}

\noindent\textbf{Problem:} Given a monocular video $\{I_t\}$ with an object of interest (indicated by a segmentation mask $\{S_t\}$), we tackle the nonrigid 3D shape and motion estimation problem, which includes estimating (1) ${\bf S}$: the rest shape of the object, (2) ${\bf D_t}$: the time-varying articulations as well as the object root body transformations, and (3) ${\bf K_t}$: the camera intrinsics.

\noindent\textbf{Overview:}
Figure~\ref{fig:overview} illustrates the overview of our method.
Motivated by recent progress in differentiable rendering and self-supervised shape learning~\cite{cmrKanazawa18,liu2019softras}, we cast the nonrigid 3D shape and motion estimation problem as an analysis-by-synthesis task. Despite the under-constrained nature of this problem, we hypothesize that, by giving appropriate video measurements, a ``low-rank'' shape and motion can be solved up to an unknown scale. %
Model parameters ${\bf X = \{S,D_t,K_t\}}$ %
are updated (via gradient descent) to minimize the difference between the rendered output ${\bf Y} \!=\! {\bf f} ({\bf X})$ and ground-truth video measurements ${\bf Y^*}$ at \emph{test} time (Sec.~\ref{sec:model}). To deal with the fundamental ambiguities in object shape, deformation and camera motion, we seek (1) a "low-rank" but expressive parameterization of deformation (Sec.~\ref{sec:deformation}), (2) rich constraints provided by optical flow and raw pixels, and (3) appropriate regularization of object shape deformation and camera motion (Sec.~\ref{sec:learning}).

\subsection{Forward-synthesis model}
\label{sec:model}
We first introduce the forward synthesis model. Given a frame index $t$ and model parameters ${\bf X}$, %
we synthesize the measurements of the corresponding frame pair $\{ t, t+1\}$, including color images renderings $\{\hat{I}_t,\hat{I}_{t+1}\}$, object silhouettes renderings $\{\hat{S}_t,\hat{S}_{t+1}\}$ and forward-backward optical flow renderings $\{{\bf \hat{u}}^{+}_t,{\bf \hat{u}}^{-}_{t+1}\}$.

\noindent\textbf{Rendering pipeline:} 
We represent object shape as a triangular mesh ${\bf S}=\{{\bf \bar{V}}, {\bf \bar{C}}, {\bf F}\}$ with vertices ${\bf \bar{V}}\in \mathbb{R}^{N\times3}$, vertex colors ${\bf \bar{C}}\in \mathbb{R}^{N\times3}$ and a fixed topology ${\bf F}\in \mathbb{R}^{M\times3}$. To model time-varying articulations ${\bf D_t}$, we have 
\begin{equation}\footnotesize
{\bf V}_{t}= {\bf G_{0,t}}({\bar {\bf V}} + \Delta{\bf V_t})\\
\end{equation}
where $\Delta{\bf V_t}$ is a per-vertex motion field applied to the rest vertices ${\bf \bar{V}}$ , and ${\bf G_{0,t}}=\begin{pmatrix}[c|c]{\bf R_{0}} & {\bf T_{0}}\end{pmatrix}{\bf_t}$ is an object root body transformation matrix (index 0 is used to differentiate from bone transformations indexed from 1 in Sec.~\ref{sec:deformation}). Finally, we apply a perspective projection ${\bf K_t}$ before rasterization, where principal point $(p_x,p_y)$ is assumed to be constant and focal length $f_t$ varies over time to deal with zoom-in/out.

\noindent\textbf{Shaders:} We render object silhouette and color images with a differentiable renderer~\cite{liu2019softras}. Color images are rendered given per-vertex appearance $\bar{\bf C}$ and constant ambient light. To synthesize the forward flow ${\bf u}^{+}_t$, we take surface positions ${\bf V}_t$ corresponding to each pixel in frame $t$, compute their locations ${\bf V}_{t+1}$ in the next frame, then take the difference of their projections:
\begin{equation}\footnotesize
    \begin{pmatrix}
    u_{x,t}^{+} \\ u_{y,t}^{+} \end{pmatrix}
    = \begin{pmatrix} {\bf P}^{(1)}_t {\bf V}_t /  {\bf P}^{(3)}_t {\bf V}_t  \\ {\bf P}^{(2)}_t {\bf V}_t /  {\bf P}^{(3)}_t {\bf V}_t \end{pmatrix} 
    -  \begin{pmatrix} {\bf P}^{(1)}_{t+1} {\bf V}_{t+1} /  {\bf P}^{(3)}_{t+1} {\bf V}_{t+1}  \\ {\bf P}^{(2)}_{t+1} {\bf V}_{t+1} /  {\bf P}^{(3)}_{t+1} {\bf V}_{t+1} \end{pmatrix},
\end{equation}
where ${\bf P}^{(i)}$ denotes the $i$th row of the projection matrix ${\bf P}$.

\subsection{Articulation Modeling} 
\label{sec:deformation}
\noindent\textbf{Unknowns vs constraints:} 
Similar to NRSfM, we analyse the number of unknowns and constraints to solve the inverse problem. Given $T$ frames of a video, we have 
\begin{equation}\footnotesize
\begin{alignedat}{4}
    \text{\# Unknowns} = & 3N \quad + \quad && 3NT \quad  +  \quad   && 6T \quad +  \quad\quad&&(T+2),\\
                         &({\bf \bar{V}}) && (\Delta{\bf V})   && ({\bf R_0}, {\bf T_0})&&({\bf K})\\
\end{alignedat}
\end{equation}
which grows linearly with the number of vertices. Motivated by NRSfM~\cite{dai2014simple} that uses low-rank shape and motion basis to deal with the exploding solution space, we seek an expressive but low-rank representation of shape and motion.

\noindent\textbf{Linear-blend skinning:}
Instead of modeling deformation as per-vertex motion $\Delta{\bf V_t}$~\cite{ucmrGoel20,cmrKanazawa18,li2020self}, we adopt a linear-blend skinning model (LBS)~\cite{kulkarni2020articulation,lewis2000pose} to constrain vertex motion by blending $B$ rigid ``bone" transformations $\{{\bf G_1},\cdots,{\bf G_B}\}$, which reduces the number of parameters and makes optimization easier. Besides bone transformations, the LBS model defines a skinning weight matrix ${\bf W}\in \mathbb{R}^{B\times N}$ that attaches the vertices of rest shape vertices ${\bf \bar{V}}$ to the set of bones. Each vertex is transformed by linearly combining the weighted bone transformations in the object coordinate frame and then transformed to the camera coordinate frame,
\begin{equation}\footnotesize
{\bf V}_{i,t}={\bf G_{0,t}}(\sum\nolimits_b {\bf W}_{b,i}{\bf G}_{b,t}){\bar {\bf V}}_{i}
\end{equation}
where $i$ is the vertex index, $b$ is the bone index.
Unlike A-CSM~\cite{kulkarni2020articulation} that only learns articulation, we learn skinning weights and time-varying bone transformations jointly.

\noindent\textbf{Parametric skinning model:} 
Inspired by the work on surface editing and local 3D shape learning~\cite{genova2020local,sumner2007embedded}, we model the skinning weights as a mixture of Gaussians with $B$ components. The probability of assigning vertex $i$ to component $b$ is given as
\begin{equation}\footnotesize
W_{b,i} = Ce^{-\frac{1}{2}({\bf v}_i-{\bf J}_b)^T{\bf Q}_b({\bf v}_i-{\bf J}_b)},
\end{equation}
where ${\bf J}_b\in \mathbb{R}^{3}$ is the center of b-th Gaussian, ${\bf Q}_b$ is the corresponding precision matrix that determines the orientation and radius of a Gaussian, and $C$ is a normalization factor that ensures the probabilities of assigning a vertex to different Gaussians sum up to one. ${\bf W \rightarrow \{ Q, J\}}$ is optimized. Note that the mixture of Gaussian models not only reduces the number of  parameters for skinning weights from $NB$ to $9B$, but also guarantees smoothness, the benefits of which are empirically validated in our experiments (Tab.~\ref{tab:aba}). The number of shape and motion parameters now becomes
\begin{equation}\footnotesize
\begin{alignedat}{5}
    \text{\# Unknowns} = & 3N \,+\, && 6BT \qquad+\quad&&9B\quad+\, &&\quad  6T \quad+  \quad&&(T+2),\\
                         &({\bf \bar{V}})&&({\bf G_{1\dots B}}) && ({\bf J},{\bf Q})  && ({\bf R_0}, {\bf T_0})&&({\bf K})\\
\end{alignedat}
\end{equation}
which grows linearly \wrt the number of frames and bones. 

\begin{figure}
    \centering
    \includegraphics[width=\linewidth]{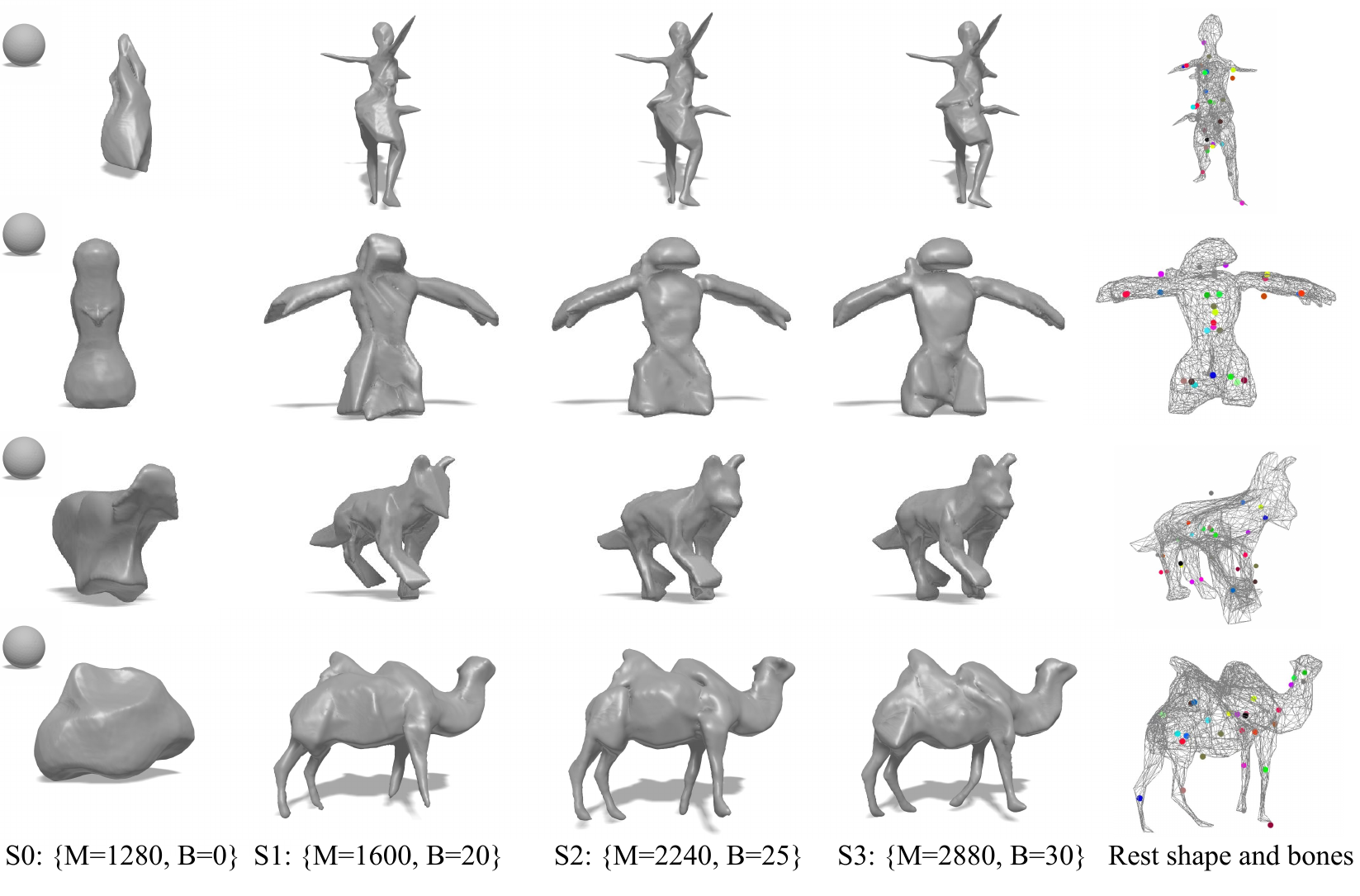}
    \caption{Coarse-to-fine reconstruction from S0 to S3. The learned centers of 3D Gaussians are shown as colored dots. }
    \label{fig:ctf}
\end{figure}

\subsection{Self-supervised Learning from a Video}
\label{sec:learning}
We exploit rich supervision signals from dense optical flow and raw pixels, as well as shape and motion regularizers to further constrain the problem.  

\noindent\textbf{Reconstruction Losses:} The supervision for our analysis-by-synthesis pipeline includes silhouette loss, optical flow loss, texture loss, and perceptual loss. Given a pair of rendered outputs $(\hat{S}_t, \hat{{I}}_t, \hat{{\bf u}}_t)$ and measurements $(S_t,{I}_t,{\bf u}_t)$, the inverse graphics loss is computed as,
\begin{equation}\footnotesize
\begin{alignedat}{3}
L_{\textrm{IG}} = &\beta_1||\hat{S}_t^i - S_t||^2_2 + \beta_2\sigma_t||\hat{{\bf u}}_t^i - {\bf u}_t||_2 &+ \beta_3||\hat{{I}}_t^i - {I}_t||_1 \\ 
 &&+\beta_4 \text{\emph{pdist}}(\hat{{I}}_t, {I}_t)
\end{alignedat}
\end{equation}
where $\{\beta_1,\cdots,\beta_4\}$ are weights empirically chosen, $\sigma_t$ is the normalized confidence map for flow measurement, and \emph{pdist}$(\cdot,\cdot)$ is the perceptual distance~\cite{zhang2018unreasonable} measured by an AlexNet pretrained on ImageNet. Applying L2 norm loss to optical flow is empirically better than squared L2 loss, and we hypothesize the reason being that the former is more tolerant to outliers in the observed flow fields.

\noindent\textbf{Shape and motion regularization:}
We exploit generic shape and temporal regularizers to constrain the problem. A Laplacian operator is applied to the rest mesh to enforce smooth surfaces,
\begin{equation}\footnotesize
\begin{alignedat}{3}
L_{\textrm{shape}} =  ||\bar{\bf V}_i -\frac{1}{|N_i|}\sum_{j\in N_i} \bar{\bf V}_j||^2.
\end{alignedat}
\end{equation}
Motion regularization includes an ARAP (as-rigid-as-possible) deformation term and a least deformation term. The ARAP term encourages natural deformation~\cite{sumner2007embedded,tulsiani2020imr},
\begin{equation}\footnotesize
\begin{alignedat}{3}
L_{\textrm{ARAP}} = \sum_{i=1}^{V}\sum_{j\in N_i}|\  || {\bf V}_{i,t}-{{\bf V}}_{j,t}  ||_2 - || {\bf V}_{i,t+1}-{\bf V}_{j,t+1}  ||_2 \ |.
\end{alignedat}
\end{equation}
The least deformation term encourages the deformation from the rest shape to be small~\cite{cmrKanazawa18},
\begin{equation}\footnotesize
\begin{alignedat}{3}
L_{\textrm{least-motion}} = \sum_{i=1}^{V} || {\bf V}_{i,t}-\bar{\bf V}_{i}  ||_2,
\end{alignedat}
\end{equation}
which discourages arbitrarily large deformations and reduces ambiguities in joint object root body pose and articulation recovery.

\noindent\textbf{Soft-symmetry constraints:}
To exploit the reflectional symmetry structure exhibited in common objects, we pose a soft-symmetry constraint along the symmetry plane $({\bf n}^*,0)$ at an arbitrary frame $t^*$. The symmetry plane is initialized from visual inspection and jointly optimized. We encourage the rest shape to be similar to its reflection,
\begin{equation}\footnotesize
\begin{alignedat}{3}
L_{\textrm{symm-shape}} = L_{\text{cham}}(\{ \bar{\bf V} ,{\bf F}\},\{{\bf H}\bar{\bf V},{\bf F} \})
\end{alignedat}
\end{equation}
where ${\bf H}={\bf I}-2{\bf n}_*{\bf n}_*^T$ is the Householder reflection matrix, and the Chamfer distance ($L_{\text{cham}}$) is computed as bidirectional pixel-to-face distances. For the centers of Gaussian control points ${\bf J}$, we also have
\begin{equation}\footnotesize
\begin{alignedat}{3}
L_{\textrm{symm-bone}} = L_{\text{cham}}( \bar{\bf J},{\bf H}\bar{\bf J}).
\end{alignedat}
\end{equation}
The total loss is a weighted sum of all losses with the weights empirically chosen and fixed for all experiments. 

\begin{figure}
    \centering
    \includegraphics[width=\linewidth]{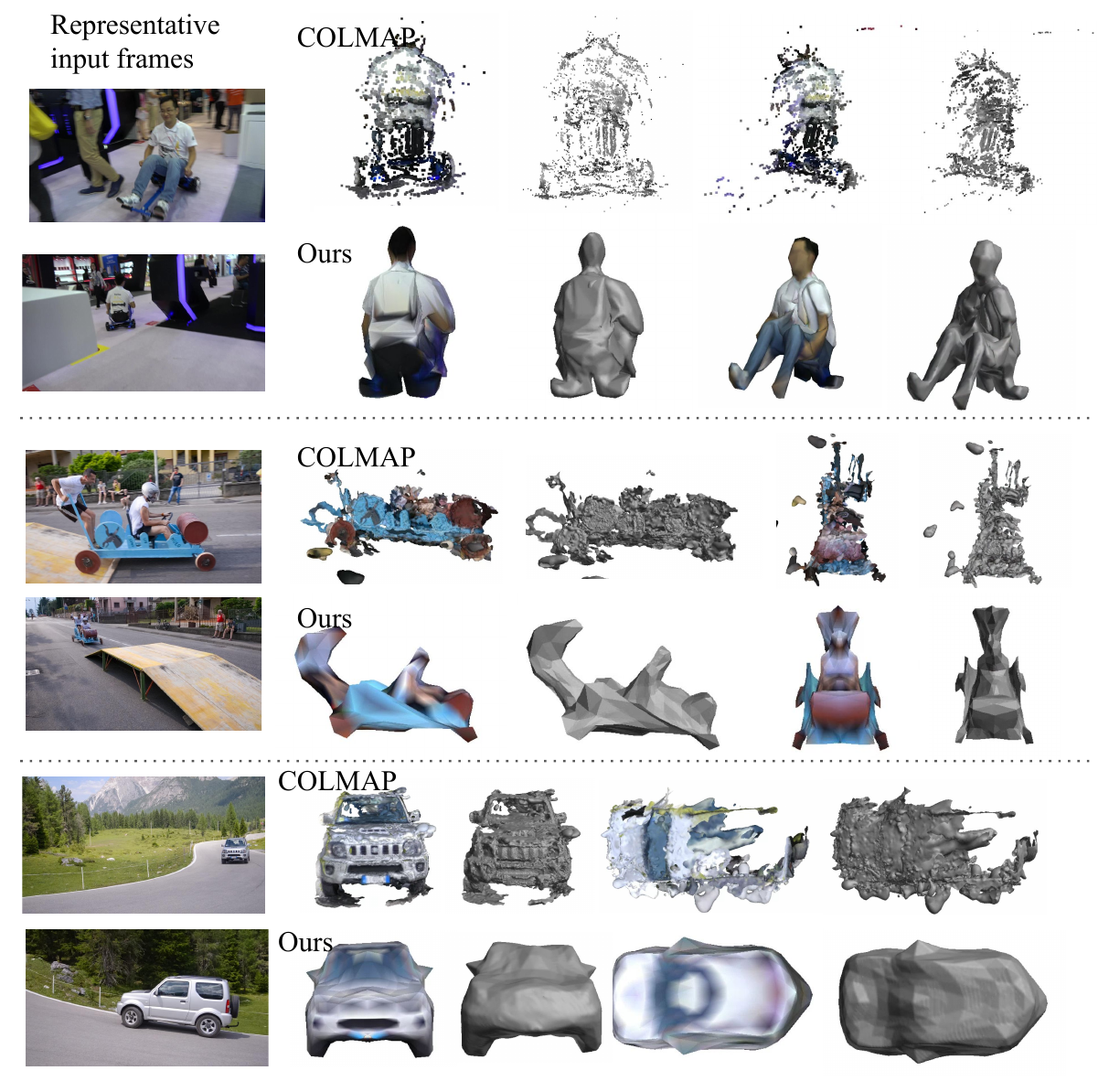}
    \caption{Visual comparison on near-rigid DAVIS sequences: scooter, soapbox, and car. COLMAP reconstructs only the visible rigid part, while our method faithfully reconstruct both the rigid object and near-rigid person.}
    \label{fig:badja-scooter}
    \vspace{-10pt}
\end{figure}

\subsection{Implementation Details}
\label{sec:details}
\noindent\textbf{Neural basis for time-varying parameters:}
Instead of optimizing explicit time-varying parameters $\{{\bf D_t}, {\bf K_t}\}$, we parameterize those as predictions from a convolutional network (ResNet-18~\cite{he2016deep}) given an input image ${\bf I}_t$,
\begin{equation}\footnotesize
\begin{alignedat}{3}
\psi_w({\bf I}_t) &= ({\bf K}, {\bf G}_{0}, {\bf G}_{1},{\bf G}_2,\cdots,{\bf G}_{B})_t,\\
\end{alignedat}
\end{equation}
where one parameter is predicted for focal length, four parameters are predicted for each bone rotation parameterized by quaternion, and three numbers are predicted for each translation, adding to $1+7(B+1)$ numbers in total at each frame. The weights are initialized 
with ImageNet~\cite{deng2009imagenet} pre-training and then optimized by LASR for each test video. 
Intuitively, the network learns a joint basis for cameras and poses that is empirically much easier to optimize than the raw parameters (Tab.~\ref{tab:aba}). 

\noindent\textbf{Silhouette and flow measurements}
Our approach assumes that a reliable segmentation of the foreground object is given, which can be manually annotated~\cite{perazzi2016benchmark}, or estimated using instance segmentation and tracking methods~\cite{kirillov2020pointrend, zhou2020motion}. Our method requires reasonable optical flow estimation, which can be provided by state-of-the-art flow estimators~\cite{teed2020raft,yang2019volumetric} trained on a mixture of datasets~\cite{rvc}. Notably, LASR recovers from some bad flow initialization and obtains better long-term correspondences (Tab.~\ref{tab:PCK}).

\noindent\textbf{Coarse-to-fine reconstruction}
We adopt a coarse-to-fine strategy to reconstruct high-quality meshes inspired by Point2Mesh~\cite{hanocka2020point2mesh}. \textbf{S0}: We first assume a rigid object and optimize the rest shape and cameras $\{{\bf S}, {\bf G_{0,t}}, {\bf K_t} \}$ for 20 epochs. The rest shape is initialized from a subdivided icosahedron projected onto a sphere. \textbf{S1-S3:} We perform iso-surface extraction and re-meshing~\cite{huang2018robust} to fix mesh self-intersections and long edges. After remeshing, the number of vertices and the number of bones increase, as shown in Fig.~\ref{fig:ctf}. The centers of Gaussian control points are initialized by running K-means on the vertices coordinates. We then jointly optimize all parameters $\{{\bf S}, {\bf D_{t}}, {\bf K_t} \}$ for 10 epochs. The above procedure is repeated three times (S1-S3).

\section{Experiments}
\noindent\textbf{Setup:}
Due to the difficulty of obtaining 3D ground truth for nonrigid objects in the real world, we evaluate 2D keypoint transfer accuracy as a proxy of 3D reconstruction quality on real videos. We additionally evaluate 3D reconstruction accuracy on objects with ground-truth meshes.

\begin{figure}
    \centering
    \includegraphics[width=\linewidth]{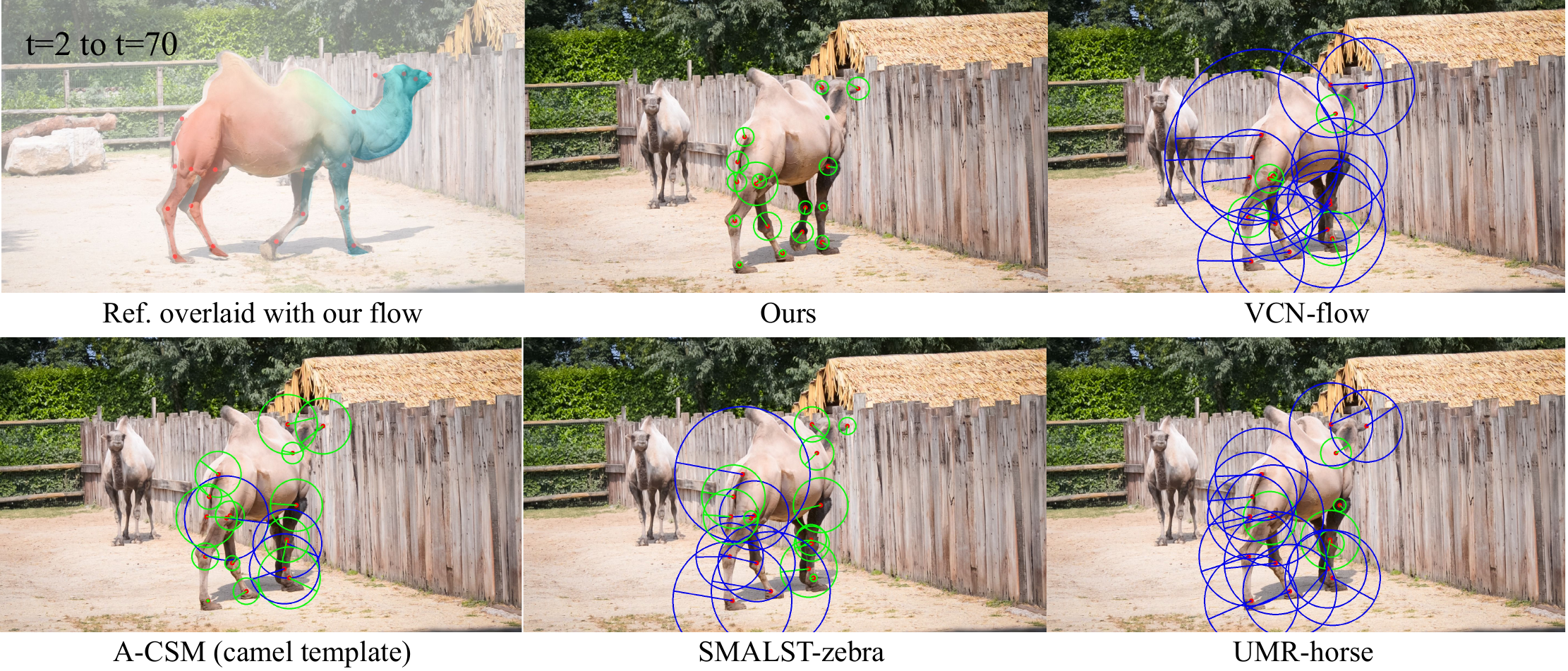}
    \caption{Keypoint transfer between frame 2 and frame 70 of the camel video. The distance between tranferred keypoint and target annotation is represented by the radius of circles. A correct transfer is marked with green and a wrong transfer is marked with blue. Our method transfers keypoint more accurately than baselines.}
    \label{fig:kp-camel}
    \vspace{-10pt}
\end{figure}

\subsection{2D Keypoint Transfer on Animal Videos}
\noindent\textbf{Dataset:} We test our method on an animal video dataset, {\bf BADJA}~\cite{biggs2018creatures}, which provides nine real animal videos with 2D keypoint and mask annotations, derived from the DAVIS video segmentation dataset~\cite{perazzi2016benchmark} and online stock footage. It includes three videos of dogs, two videos of horsejump, and one video of camel, cow, bear as well as impala. We report quantitative results on one video per-category and show the reconstruction of the rest in the sup. mat.

\noindent\textbf{Metric:} To approximate the accuracy of 3D shape and articulation recovery, we adopt {\bf percentage of correct keypoint transfer} (PCK-T)~\cite{cmrKanazawa18,kulkarni2020articulation,yang2011articulated} metric. Given a reference and target image pair with 2D keypoint annotations, the reference keypoint is transferred to the target image, and labeled as ``correct'' if the transferred keypoint is within some threshold distance $d_{th}=0.2\sqrt{|S|}$ from the target keypoint, where $|S|$ is the area of the ground-truth silhouette~\cite{biggs2018creatures}. In practice, we transfer points by re-projection from the reference frame to the target frame given the articulated shape and camera pose estimations. If the back-projected keypoint lies outside the reconstructed mesh, we re-project its nearest neighbor that intersects the mesh. The accuracy is averaged over all T(T-1) pairs of frames.

\noindent\textbf{Baselines:} We compare with state-of-the-art methods for animal reconstruction and refer to Tab.~\ref{tab:taxonomy} for a taxonomy. {\bf SMALST}~\cite{zuffi2019three} is a model-based regressor trained for zebras. It takes an image as input and predicts shape, pose and texture for the SMAL~\cite{Zuffi:CVPR:2017} model. {\bf UMR}~\cite{li2020self} is a category-specific shape estimator trained for several categories, including birds, horses and other categories that have a large collection of annotated images. We report the performance of the horse model since the models of other animal categories are not available. \textbf{A-CSM}~\cite{kulkarni2020articulation} learns a category-specific canonical surface mapping and articulations from an image collection. At test time, it takes an image as input and predicts the articulation parameters of a rigged template mesh. It provides 3D templates for 27 animal categories and an articulation model for horses, which is used throughout the experiments. {\bf SMALify}~\cite{biggs2018creatures} is a model-based optimization approach that fits one of five categories (including cat, dog, horse, cow and hippo) of SMAL models to a video or a single image. We provide all the video frames with ground-truth keypoint and mask annotations.
Close to our setup, {\bf N-NRSfM}~\cite{sidhu2020neural} trains a video-specific model for object shape, deformation and camera parameters from multi-frame optical flow estimations~\cite{garg2013variational}. Finally, we include a detection-based method, {\bf OJA}~\cite{biggs2018creatures}, which trains an hourglass network to detect animal keypoints (indicated by Detector), and post-process the joint cost maps with a proposed optimal assignment algorithm. The results of PCK are taken from the paper~\cite{biggs2018creatures} without recomputing PCK-T.

\begin{table}[!t]
    \caption{2D Keypoint transfer accuracy on BADJA.  $^\text{(1)}$Model-based regression.  $^\text{(2)}$Category-specific reconstruction.  $^\text{(3)}$Free-form reconstruction. Methods with$^{\dagger}$ do not reconstruct 3D shape. Results with$^{*}$ indicates the method is not designed for such category. Best results are underlined, and bolded if reconstruct a 3D shape. %
    }
    \small
    \centering
    \begin{tabular}{rlllllll}
	\toprule
 Method                                         & camel & dog   & cows  & horse & bear\\
 \midrule
 $^\text{(1)}$SMALST~\cite{zuffi2019three}                &49.7$^{*}$ & 12.8$^{*}$ & 59.7$^{*}$ & 10.4$^{*}$ & 67.2$^{*}$\\
 $^\text{(2)}$A-CSM~\cite{kulkarni2020articulation}          &60.2$^{*}$ & 24.5$^{*}$ & 65.7$^{*}$ & 21.5 & 39.7$^{*}$\\
 $^\text{(2)}$UMR~\cite{li2020self}                          &35.1$^{*}$ & 38.5$^{*}$ & 68.1$^{*}$ & 32.4 & 56.9$^{*}$ \\
 $^\text{(3)}$N-NRSfM~\cite{sidhu2020neural} & 67.8 & 17.9  &70.0& 8.7 &60.2\\
 $^\text{(3)}$LASR (Ours)                        & {\bf 81.9}  & {\bf 65.8}&{\bf 83.7}  & \underline{{\bf 49.3}} &{\bf 85.1}\\
 $^\text{(3)}$ +Auto-mask                                      &78.9 &59.5&82.7&42.2&82.6\\
\midrule
 $^{\dagger}$Static                                        & 51.9  & 13.0 & 55.5 & 8.8  &58.6\\
 $^{\dagger}$Detector~\cite{biggs2018creatures} &73.3   & 66.9  & 89.2  & 26.5  & 83.1\\ 
 $^{\dagger}$OJA~\cite{biggs2018creatures}      &\underline{87.1}   & \underline{66.9}  & \underline{94.7}  & 24.4  & \underline{88.9}\\
 $^{\dagger}$Flow-VCN~\cite{yang2019volumetric}  & 47.9  & 25.7 & 60.7 & 14.4 &63.8\\
 \bottomrule
\label{tab:PCK}
\end{tabular}
\vspace{-20pt}
\end{table}

\begin{figure*}[!ht]
    \centering
    \includegraphics[width=\linewidth]{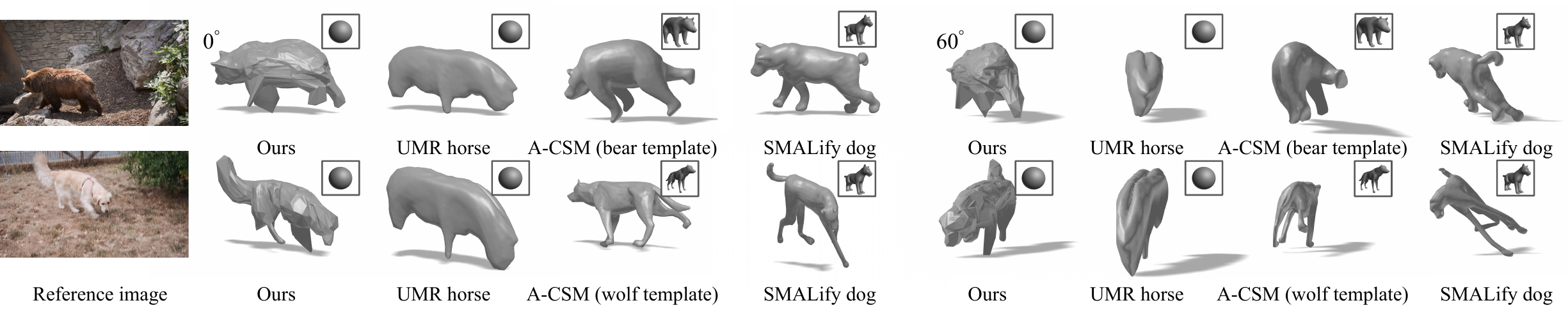}
    \caption{Comparison on BADJA bear and dog videos. The reconstruction of the first frame of the video is shown from two viewpoints. Compared to UMR that also does not use a shape template, LASR reconstructs more fine-grained geometry. Compared to A-CSM and SMALify that use a shape template, LASR recovers instance-specific details, such as the fluffy tail of the dog, and a more natural pose.}
    \label{fig:badja-animals}
\end{figure*}

\noindent\textbf{Results:} Qualitative results of 3D shape reconstruction are shown in Fig.~\ref{fig:teaser} and Fig.~\ref{fig:badja-animals}, where we compare with UMR, A-CSM and SMALify on the camel, bear and dog video.  Quantitative results of keypoint transfer are shown in Tab.~\ref{tab:PCK}. Given that all 3D reconstruction baselines are category-specific and might not provide the exact model for some categories (such as camel), we pick up the best model or template for each animal video. Compared with 3D reconstruction baselines, LASR is better for all categories, even on the categories the baselines are trained for (e.g., LASR: 49.3 vs UMR: 32.4 on horsejump-high). Replacing the GT masks with an object segmentor, PointRend~\cite{kirillov2020pointrend}, the performance of LASR (`+Auto-mask' in Tab.~\ref{tab:PCK}) drops, but is still better than all the reconstruction baselines. Compared to detection-based methods, our accuracy is higher on the horsejump video, and close to the baseline on other videos. LASR also shows a large improvement compared to the initial optical flow (81.9\% vs 47.9\% for camel), especially between long-range frames as shown in Fig.~\ref{fig:kp-camel}. 

\begin{table}[!t]
    \caption{Mesh reconstruction error in terms of Chamfer distance on our animated object dataset. To ensure comparable results over objects, ground-truth shapes are rescaled such that the maximum distance between vertices is 10.  Best results are bolded. ``-'' means a method does not apply to a particular sequence. }
    \small
    \centering
    \begin{tabular}{rcccccccc}
	\toprule
 Method  & dancer $\downarrow$ & dog $\downarrow$ & horse $\downarrow$ & golem $\downarrow$ \\
 \midrule
 SMPLify-X~\cite{SMPL-X:2019}            & 0.26 &-&-&-\\
 VIBE~\cite{kocabas2019vibe}             & {\bf 0.22} &-&-&-\\
  A-CSM~\cite{kulkarni2020articulation}  &-&0.38&0.26&-\\
 SMALify~\cite{biggs2018creatures}      &-& 0.51 & 0.41 &-\\
 PIFuHD~\cite{saito2020pifuhd}      & 0.28&-&-&-\\
  \midrule
  UMR~\cite{li2020self}             &-    & 0.44 &0.42 &-\\
 LASR (Ours)                               & 0.35  & {\bf 0.28}&{\bf 0.23}&0.16\\
 \bottomrule
\label{tab:mesh}
\end{tabular}
\vspace{-20pt}
\end{table}

\subsection{Mesh Reconstruction on Articulated Objects}

\noindent\textbf{Dataset:}
To evaluate mesh reconstruction accuracy, we collect a video dataset of five articulated objects with ground-truth mesh and articulation, including one dancer video from AMA (Articulated Mesh Animation dataset)~\cite{vlasic2008articulated}, one German shepherd video, one horse video, one eagle video and one stone golem video from TurboSquid. We also include a rigid object, Keenan's spot to evaluate performance on rigid object reconstruction and ablation for the S0 stage. 

\noindent\textbf{Metric:} Most prior work on mesh reconstruction assumes given camera parameters. However, both the camera and the geometry are unknown in our case, which leads to ambiguities in evaluation, including scale ambiguity (exists for all monocular reconstruction) as well as the depth ambiguity (exists for weak perspective cameras as used in UMR, A-CSM, VIBE, etc.). To factorize out the unknown camera matrices, we align two meshes with a 3D similarity transformation solved by ICP. Then, the bidirectional Chamfer distance is adopted as the evaluation metric. We follow prior work~\cite{meshrcnn,ravi2020pytorch3d} to randomly sample 10k points uniformly from the surface of predicted and ground-truth meshes, and compute the average distance between the nearest neighbor for each point in the corresponding point cloud.

\noindent\textbf{Baselines:} Besides A-CSM, SMALify, and UMR for animal reconstruction, we compare with SMPLify-X, VIBE, and PiFUHD for human reconstruction. 
{\bf SMPLify-X}~\cite{SMPL-X:2019} is a model-based optimization method for expressive human body capture. We use the female SMPL model for the dancer sequence, and provide the keypoint inputs estimated from OpenPose~\cite{8765346}. {\bf VIBE}~\cite{kocabas2019vibe} is a state-of-the-art model-based video regressor for human pose and shape inference. {\bf PIFuHD} is a state-of-the-art free-form 3D shape estimator for clothed humans. It takes a single image as input and predicts an implicit shape representation, which is converted to a mesh by the marching cube algorithm. To compare with SMALify on dog and horse, we manually annotate 18 keypoints per-frame, and initialize with the 
corresponding shape template.

\begin{table}[!t]
    \caption{
Ablation study with mesh reconstruction error.}
    \small
    \centering
    \begin{tabular}{lcccccccc}
	\toprule
 {\bf S0} & ref. & $^\text{(1)}$w/o flow & $^\text{(2)}$w/o L$_{can}$ & $^\text{(3)}$w/o CNN  \\
\midrule
 spot   & 0.05 & 0.55   & 0.61 & 0.63\\
\midrule
 {\bf S0-S3} & ref. & $^\text{(4)}$w/o LBS & $^\text{(5)}$w/o C2F & $^\text{(6)}$w/o GMM \\
\midrule
 dog  & 0.28 & 0.68 & 0.59 &0.34\\
 \bottomrule
\label{tab:aba}
\end{tabular}
\vspace{-20pt}
\end{table}

\begin{figure*}[!ht]
    \centering
    \includegraphics[width=\linewidth]{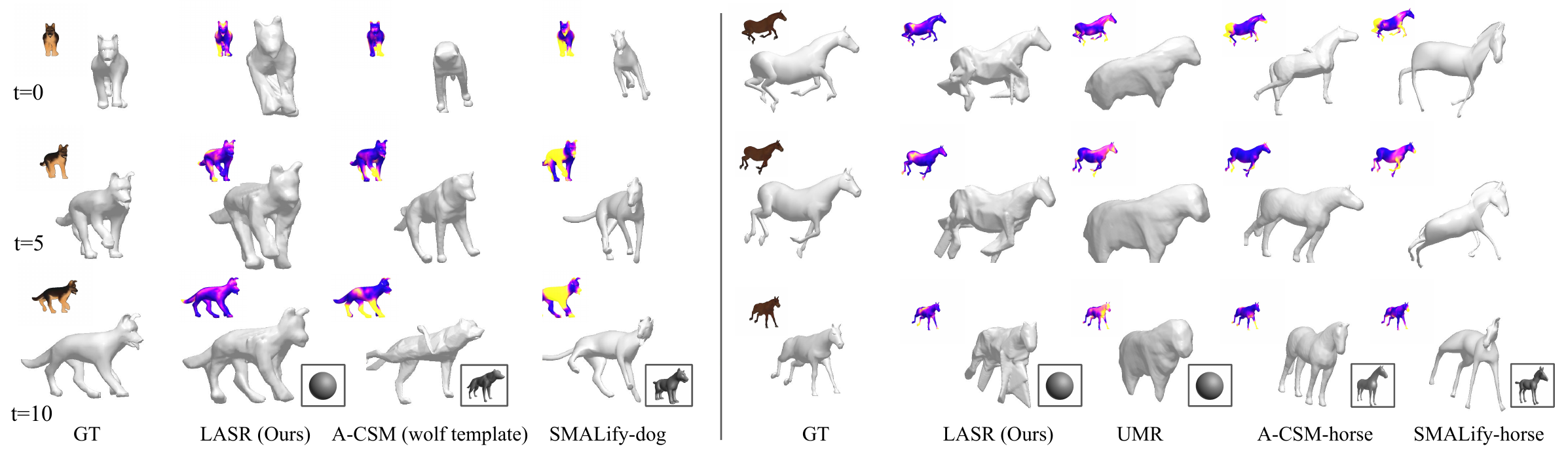}
    \caption{Shape and articulation reconstruction results on synthetic dog and horse sequences. We also visualize Chamfer distances measured from the ground truth to the reconstructed mesh on top of each result, and yellow indicates high error. Compared to UMR, LASR successfully reconstructs the four legs of the horse. Compared to template-based methods (A-CSM and SMALify), LASR successfully reconstructs the instance-specific details (ears and tails of the dog) and recovers a more natural articulation. The reference is shown at the left corner and the template mesh used is shown in the bottom right boxes.}
    \label{fig:sdog}
\end{figure*}

\noindent{\bf Results} The visual comparison on human and animals are shown in Fig.~\ref{fig:teaser} and Fig.~\ref{fig:sdog} respectively. We report the quantitative results in Tab.~\ref{tab:mesh}. On the dog video, our method is better than all the baselines (0.28 vs A-CSM: 0.38), possibly because A-CSM and UMR are not trained specifically for dogs (although A-CSM uses a wolf template), and SMALify cannot reconstruct a natural 3D shape from limited keypoint and silhouette annotations. For the horse video, our method is slightly better than A-CSM, which uses a horse shape template, and outperforms other baselines. For the dancer sequence, our method is not as accurate as baseline methods (0.35 vs VIBE: 0.22), which is expected given that all baselines either use a well-designed human model, or have been trained with 3D human mesh data, while LASR does not have access to 3D human data. For the stone golem video, our method is the only one that reconstructs a meaningful shape. Although the stone golem has a similar shape to a human's, OpenPose does not detect joints correctly, leading to the failure of SMALify-X, VIBE and PiFUHD.

\begin{figure*}[!ht]
    \centering
    \includegraphics[width=\linewidth]{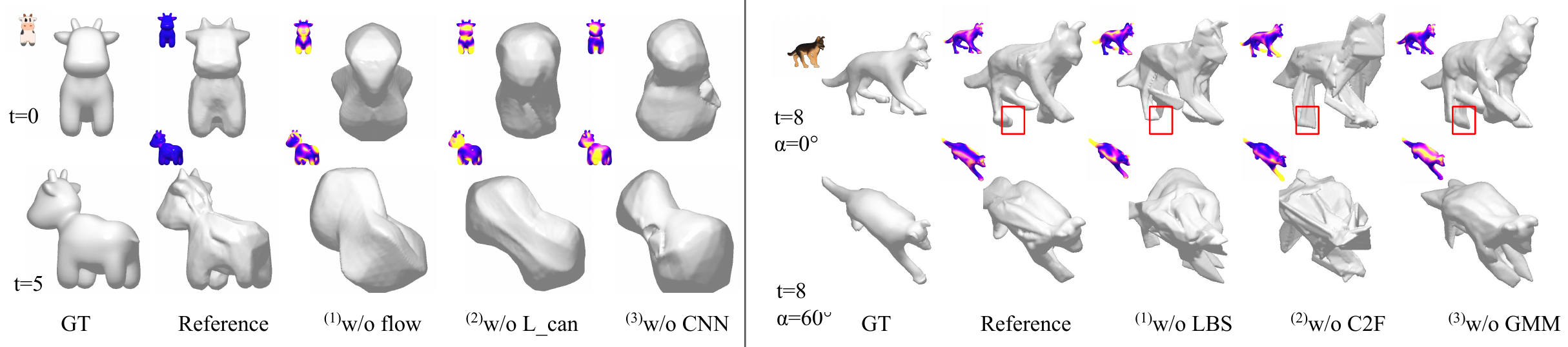}
    \caption{Left: Ablation study on camera and rigid shape optimization ({\bf S0}). Removing the optical flow loss introduces large errors in camera pose estimation and therefore the overall geometry is not recovered. Removing the canonicalization loss leads to worse camera pose estimation, and therefore the symmetric shape constraint is not correctly enforced.  Finally, if we directly optimize the camera poses without using a convolutional network, it converges much slower and does not yield an ideal shape within the same iterations. Right: Ablation study on articulated shape optimization ({\bf S1-S3}). We show the reconstructed articulated shape at the middle frame (t=8) from two viewpoints. Without LBS model, although the reconstruction looks plausible from the visible view, it does not recover the full geometry due to the redundant deformation parameters and lack of constraints. Without coarse-to-fine re-meshing, fine-grained details are not recovered. Replacing GMM skinning weights (9xB parameters) with an NxB matrix leads to extra limbs and tails on the reconstruction.}
    \label{fig:aba}
    \vspace{-10pt}
\end{figure*}

\noindent{\bf Qualitative results on DAVIS videos:}
\noindent To examine the performance on arbitrary real-world objects, we use five DAVIS videos, including dance-twirl, scooter-board, soapbox, car-turn, mallard-fly, and a cat video captured by us and segmented by PointRend. The comparison with COLMAP~\cite{schoenberger2016vote}, a template-free SfM-MVS pipeline, is shown in Fig.~\ref{fig:badja-scooter}. More results are available in the sup. mat.

\noindent{\bf Ablation study:} We investigate the effect of different design choices on the rigid ``spot'' and animated dog sequences. The videos are rendered into T=15 frames given ambient light and a camera rotating around the object by 90 degrees at zero elevation. Besides color images, we render silhouette and optical flow as the supervision. Results are shown in Fig.~\ref{fig:aba} and quantitative results are reported in Tab.~\ref{tab:aba}. In terms of camera parameter optimization and rigid shape reconstruction ({\bf S0}), we find it beneficial to use $^{(1)}$optical flow as supervision signals, $^{(2)}$canonicalization of symmetry plane, and $^{(3)}$CNN as an implicit representation for camera parameters. For articulated shape reconstruction ({\bf S1-S3}), it is critical to use $^{(4)}$linear blend skinning, $^{(5)}$coarse-to-fine re-meshing, and $^{(6)}$parametric skinning model.

\vspace{1mm}
\noindent{\bf Limitations:} %
Empirically, LASR struggles to estimate surfaces that are not visible in any input view and fails at heavy occlusions that are missed by mask annotations. Its efficiency also needs improvement, as it takes less than one hour for rigid objects and a few hours for nonrigid shapes on a single GPU. %

\section{Conclusion}
We present LASR, a template-free approach for articulated shape reconstruction from a monocular video. LASR faithfully reconstructs individual objects from diverse categories (such as human, camel, dog, bear, etc.) without relying on category-specific shape templates, making it applicable to a wide range of scenarios.
We hope that LASR will enable more progress in articulated shape reconstruction.

\newpage

{\small
\bibliographystyle{ieee_fullname}
\bibliography{egbib}

\begin{thebibliography}{10}\itemsep=-1pt

\bibitem{rvc}
Robust vision challenge.
\newblock \url{www.robustvision.net}, 2020.

\bibitem{agudo2018image}
Antonio Agudo, Melcior Pijoan, and Francesc Moreno-Noguer.
\newblock Image collection pop-up: 3d reconstruction and clustering of rigid
  and non-rigid categories.
\newblock In {\em CVPR}, pages 2607--2615, 2018.

\bibitem{badger2020}
Marc Badger, Yufu Wang, Adarsh Modh, Ammon Perkes, Nikos Kolotouros, Bernd
  Pfrommer, Marc Schmidt, and Kostas Daniilidis.
\newblock {3D} bird reconstruction: a dataset, model, and shape recovery from a
  single view.
\newblock In {\em ECCV}, 2020.

\bibitem{baker2005shape}
Simon Baker, Takeo Kanade, et~al.
\newblock Shape-from-silhouette across time part i: Theory and algorithms.
\newblock {\em IJCV}, 62(3):221--247, 2005.

\bibitem{biggs2020wldo}
Benjamin Biggs, Ollie Boyne, James Charles, Andrew Fitzgibbon, and Roberto
  Cipolla.
\newblock {W}ho left the dogs out: {3D} animal reconstruction with expectation
  maximization in the loop.
\newblock In {\em ECCV}, 2020.

\bibitem{biggs2018creatures}
Benjamin Biggs, Thomas Roddick, Andrew Fitzgibbon, and Roberto Cipolla.
\newblock Creatures great and smal: Recovering the shape and motion of animals
  from video.
\newblock In {\em ACCV}, pages 3--19. Springer, 2018.

\bibitem{Bogo:ECCV:2016}
Federica Bogo, Angjoo Kanazawa, Christoph Lassner, Peter Gehler, Javier Romero,
  and Michael~J. Black.
\newblock Keep it {SMPL}: Automatic estimation of {3D} human pose and shape
  from a single image.
\newblock In {\em ECCV}, 2016.

\bibitem{bregler2000recovering}
Christoph Bregler, Aaron Hertzmann, and Henning Biermann.
\newblock Recovering non-rigid 3d shape from image streams.
\newblock In {\em CVPR}, volume~2, pages 690--696. IEEE, 2000.

\bibitem{8765346}
Z. {Cao}, G. {Hidalgo Martinez}, T. {Simon}, S. {Wei}, and Y.~A. {Sheikh}.
\newblock Openpose: Realtime multi-person 2d pose estimation using part
  affinity fields.
\newblock {\em TPAMI}, 2019.

\bibitem{cashman2012shape}
Thomas~J Cashman and Andrew~W Fitzgibbon.
\newblock What shape are dolphins? building 3d morphable models from 2d images.
\newblock {\em PAMI}, 35(1):232--244, 2012.

\bibitem{choy_nips16}
Christopher~B Choy, JunYoung Gwak, Silvio Savarese, and Manmohan Chandraker.
\newblock Universal correspondence network.
\newblock In {\em NeurIPS}. 2016.

\bibitem{dai2014simple}
Yuchao Dai, Hongdong Li, and Mingyi He.
\newblock A simple prior-free method for non-rigid structure-from-motion
  factorization.
\newblock {\em IJCV}, 107(2):101--122, 2014.

\bibitem{deng2009imagenet}
Jia Deng, Wei Dong, Richard Socher, Li-Jia Li, Kai Li, and Li Fei-Fei.
\newblock Imagenet: A large-scale hierarchical image database.
\newblock In {\em CVPR}, pages 248--255. Ieee, 2009.

\bibitem{garg2013variational}
Ravi Garg, Anastasios Roussos, and Lourdes Agapito.
\newblock A variational approach to video registration with subspace
  constraints.
\newblock {\em IJCV}, 2013.

\bibitem{genova2020local}
Kyle Genova, Forrester Cole, Avneesh Sud, Aaron Sarna, and Thomas Funkhouser.
\newblock Local deep implicit functions for 3d shape.
\newblock In {\em CVPR}, pages 4857--4866, 2020.

\bibitem{meshrcnn}
Justin~Johnson Georgia~Gkioxari, Jitendra~Malik.
\newblock Mesh r-cnn.
\newblock {\em ICCV}, 2019.

\bibitem{gkioxari2019mesh}
Georgia Gkioxari, Jitendra Malik, and Justin Johnson.
\newblock Mesh r-cnn.
\newblock In {\em ICCV}, pages 9785--9795, 2019.

\bibitem{ucmrGoel20}
Shubham Goel, Angjoo Kanazawa, and Jitendra Malik.
\newblock Shape and viewpoints without keypoints.
\newblock In {\em ECCV}, 2020.

\bibitem{gotardo2011non}
Paulo~FU Gotardo and Aleix~M Martinez.
\newblock Non-rigid structure from motion with complementary rank-3 spaces.
\newblock In {\em CVPR}, pages 3065--3072. IEEE, 2011.

\bibitem{hanocka2020point2mesh}
Rana Hanocka, Gal Metzer, Raja Giryes, and Daniel Cohen-Or.
\newblock Point2mesh: A self-prior for deformable meshes.
\newblock {\em arXiv preprint arXiv:2005.11084}, 2020.

\bibitem{he2016deep}
Kaiming He, Xiangyu Zhang, Shaoqing Ren, and Jian Sun.
\newblock Deep residual learning for image recognition.
\newblock In {\em CVPR}, pages 770--778, 2016.

\bibitem{huang2018robust}
Jingwei Huang, Hao Su, and Leonidas Guibas.
\newblock Robust watertight manifold surface generation method for shapenet
  models.
\newblock {\em arXiv preprint arXiv:1802.01698}, 2018.

\bibitem{cmrKanazawa18}
Angjoo Kanazawa, Shubham Tulsiani, Alexei~A. Efros, and Jitendra Malik.
\newblock Learning category-specific mesh reconstruction from image
  collections.
\newblock In {\em ECCV}, 2018.

\bibitem{kingma2014adam}
Diederik~P Kingma and Jimmy Ba.
\newblock Adam: A method for stochastic optimization.
\newblock {\em arXiv preprint arXiv:1412.6980}, 2014.

\bibitem{kirillov2020pointrend}
Alexander Kirillov, Yuxin Wu, Kaiming He, and Ross Girshick.
\newblock Pointrend: Image segmentation as rendering.
\newblock In {\em CVPR}, pages 9799--9808, 2020.

\bibitem{kocabas2019vibe}
Muhammed Kocabas, Nikos Athanasiou, and Michael~J. Black.
\newblock Vibe: Video inference for human body pose and shape estimation.
\newblock In {\em CVPR}, June 2020.

\bibitem{kong2019deep}
Chen Kong and Simon Lucey.
\newblock Deep non-rigid structure from motion.
\newblock In {\em ICCV}, pages 1558--1567, 2019.

\bibitem{kulkarni2020articulation}
Nilesh Kulkarni, Abhinav Gupta, David~F Fouhey, and Shubham Tulsiani.
\newblock Articulation-aware canonical surface mapping.
\newblock In {\em CVPR}, pages 452--461, 2020.

\bibitem{lewis2000pose}
John~P Lewis, Matt Cordner, and Nickson Fong.
\newblock Pose space deformation: a unified approach to shape interpolation and
  skeleton-driven deformation.
\newblock In {\em Proceedings of the 27th annual conference on Computer
  graphics and interactive techniques}, pages 165--172, 2000.

\bibitem{vmr2020}
Xueting Li, Sifei Liu, Shalini De~Mello, Kihwan Kim, Xiaolong Wang, Ming-Hsuan
  Yang, and Jan Kautz.
\newblock Online adaptation for consistent mesh reconstruction in the wild.
\newblock In {\em NeurIPS}, 2020.

\bibitem{li2020self}
Xueting Li, Sifei Liu, Kihwan Kim, Shalini De~Mello, Varun Jampani, Ming-Hsuan
  Yang, and Jan Kautz.
\newblock Self-supervised single-view 3d reconstruction via semantic
  consistency.
\newblock {\em ECCV}, 2020.

\bibitem{lin2014microsoft}
Tsung-Yi Lin, Michael Maire, Serge Belongie, James Hays, Pietro Perona, Deva
  Ramanan, Piotr Doll{\'a}r, and C~Lawrence Zitnick.
\newblock Microsoft coco: Common objects in context.
\newblock In {\em ECCV}, pages 740--755. Springer, 2014.

\bibitem{liu2019softras}
Shichen Liu, Tianye Li, Weikai Chen, and Hao Li.
\newblock Soft rasterizer: A differentiable renderer for image-based 3d
  reasoning.
\newblock {\em ICCV}, Oct 2019.

\bibitem{SMPL:2015}
Matthew Loper, Naureen Mahmood, Javier Romero, Gerard Pons-Moll, and Michael~J.
  Black.
\newblock {SMPL}: A skinned multi-person linear model.
\newblock {\em ACM Trans. Graphics (Proc. SIGGRAPH Asia)}, 34(6):248:1--248:16,
  Oct. 2015.

\bibitem{novotny2019c3dpo}
David Novotny, Nikhila Ravi, Benjamin Graham, Natalia Neverova, and Andrea
  Vedaldi.
\newblock C3dpo: Canonical 3d pose networks for non-rigid structure from
  motion.
\newblock In {\em ICCV}, pages 7688--7697, 2019.

\bibitem{SMPL-X:2019}
Georgios Pavlakos, Vasileios Choutas, Nima Ghorbani, Timo Bolkart, Ahmed A.~A.
  Osman, Dimitrios Tzionas, and Michael~J. Black.
\newblock Expressive body capture: 3d hands, face, and body from a single
  image.
\newblock In {\em CVPR}, 2019.

\bibitem{perazzi2016benchmark}
Federico Perazzi, Jordi Pont-Tuset, Brian McWilliams, Luc Van~Gool, Markus
  Gross, and Alexander Sorkine-Hornung.
\newblock A benchmark dataset and evaluation methodology for video object
  segmentation.
\newblock In {\em CVPR}, pages 724--732, 2016.

\bibitem{ravi2020pytorch3d}
Nikhila Ravi, Jeremy Reizenstein, David Novotny, Taylor Gordon, Wan-Yen Lo,
  Justin Johnson, and Georgia Gkioxari.
\newblock Accelerating 3d deep learning with pytorch3d.
\newblock {\em arXiv:2007.08501}, 2020.

\bibitem{Reinert:2016:10.20380/GI2016.17}
Bernhard Reinert, Tobias Ritschel, and Hans-Peter Seidel.
\newblock Animated 3d creatures from single-view video by skeletal sketching.
\newblock In {\em Proceedings of Graphics Interface 2016}, GI 2016, pages
  133--141. Canadian Human-Computer Communications Society / Soci{\'e}t{\'e}
  canadienne du dialogue humain-machine, 2016.

\bibitem{NEURIPS2018_8f7d807e}
Ignacio Rocco, Mircea Cimpoi, Relja Arandjelovi\'{c}, Akihiko Torii, Tomas
  Pajdla, and Josef Sivic.
\newblock Neighbourhood consensus networks.
\newblock In {\em NeurIPS}, 2018.

\bibitem{saito2019pifu}
Shunsuke Saito, Zeng Huang, Ryota Natsume, Shigeo Morishima, Angjoo Kanazawa,
  and Hao Li.
\newblock Pifu: Pixel-aligned implicit function for high-resolution clothed
  human digitization.
\newblock In {\em ICCV}, pages 2304--2314, 2019.

\bibitem{saito2020pifuhd}
Shunsuke Saito, Tomas Simon, Jason Saragih, and Hanbyul Joo.
\newblock Pifuhd: Multi-level pixel-aligned implicit function for
  high-resolution 3d human digitization.
\newblock In {\em CVPR}, 2020.

\bibitem{sand2008particle}
Peter Sand and Seth Teller.
\newblock Particle video: Long-range motion estimation using point
  trajectories.
\newblock {\em IJCV}, 80(1):72, 2008.

\bibitem{schoenberger2016vote}
Johannes~Lutz Sch\"{o}nberger, True Price, Torsten Sattler, Jan-Michael Frahm,
  and Marc Pollefeys.
\newblock A vote-and-verify strategy for fast spatial verification in image
  retrieval.
\newblock In {\em ACCV}, 2016.

\bibitem{sidhu2020neural}
Vikramjit Sidhu, Edgar Tretschk, Vladislav Golyanik, Antonio Agudo, and
  Christian Theobalt.
\newblock Neural dense non-rigid structure from motion with latent space
  constraints.
\newblock In {\em ECCV}, pages 204--222. Springer, 2020.

\bibitem{sumner2007embedded}
Robert~W Sumner, Johannes Schmid, and Mark Pauly.
\newblock Embedded deformation for shape manipulation.
\newblock In {\em ACM SIGGRAPH 2007 papers}, pages 80--es. 2007.

\bibitem{sundaram2010dense}
Narayanan Sundaram, Thomas Brox, and Kurt Keutzer.
\newblock Dense point trajectories by gpu-accelerated large displacement
  optical flow.
\newblock In {\em ECCV}, pages 438--451. Springer, 2010.

\bibitem{teed2020raft}
Zachary Teed and Jia Deng.
\newblock Raft: Recurrent all-pairs field transforms for optical flow.
\newblock 2020.

\bibitem{tulsiani2020imr}
Shubham Tulsiani, Nilesh Kulkarni, and Abhinav Gupta.
\newblock Implicit mesh reconstruction from unannotated image collections,
  2020.

\bibitem{vlasic2008articulated}
Daniel Vlasic, Ilya Baran, Wojciech Matusik, and Jovan Popovi{\'c}.
\newblock Articulated mesh animation from multi-view silhouettes.
\newblock In {\em SIGGRAPH 2008}, pages 1--9. 2008.

\bibitem{wang2020deep}
Chaoyang Wang, Chen-Hsuan Lin, and Simon Lucey.
\newblock Deep nrsfm++: Towards 3d reconstruction in the wild.
\newblock {\em arXiv preprint arXiv:2001.10090}, 2020.

\bibitem{xiang2019monocular}
Donglai Xiang, Hanbyul Joo, and Yaser Sheikh.
\newblock Monocular total capture: Posing face, body, and hands in the wild.
\newblock In {\em CVPR}, pages 10965--10974, 2019.

\bibitem{yang2019volumetric}
Gengshan Yang and Deva Ramanan.
\newblock Volumetric correspondence networks for optical flow.
\newblock In {\em NeurIPS}, pages 794--805, 2019.

\bibitem{yang2011articulated}
Yi Yang and Deva Ramanan.
\newblock Articulated pose estimation with flexible mixtures-of-parts.
\newblock In {\em CVPR}, pages 1385--1392. IEEE, 2011.

\bibitem{zhang2018unreasonable}
Richard Zhang, Phillip Isola, Alexei~A Efros, Eli Shechtman, and Oliver Wang.
\newblock The unreasonable effectiveness of deep features as a perceptual
  metric.
\newblock In {\em CVPR}, pages 586--595, 2018.

\bibitem{zhou2020motion}
Tianfei Zhou, Shunzhou Wang, Yi Zhou, Yazhou Yao, Jianwu Li, and Ling Shao.
\newblock Motion-attentive transition for zero-shot video object segmentation.
\newblock In {\em AAAI}, pages 13066--13073, 2020.

\bibitem{zuffi2019three}
Silvia Zuffi, Angjoo Kanazawa, Tanya Berger-Wolf, and Michael Black.
\newblock Three-d safari: Learning to estimate zebra pose, shape, and texture
  from images “in the wild”.
\newblock In {\em ICCV}, pages 5358--5367. IEEE.

\bibitem{Zuffi:CVPR:2018}
Silvia Zuffi, Angjoo Kanazawa, and Michael~J. Black.
\newblock Lions and tigers and bears: Capturing non-rigid, {3D}, articulated
  shape from images.
\newblock In {\em CVPR}, 2018.

\bibitem{Zuffi:CVPR:2017}
Silvia Zuffi, Angjoo Kanazawa, David Jacobs, and Michael~J. Black.
\newblock {3D} menagerie: Modeling the {3D} shape and pose of animals.
\newblock In {\em CVPR}, July 2017.

\end{thebibliography}
}

\clearpage
\newpage

\section{Appendix}
\subsection{Implementation details} 
\noindent {\bf Training details:} We include details of the hyper-parameters used for training in Tab.~\ref{tab:hp}. 

\noindent {\bf Video pre-processing:}
We provide details for video pre-processing. To ensure enough object motion between adjacent frames, we use a heuristic rule that skips the next frame when the average magnitude of measured flow within the object silhouette is lower than 0.05 in the clip space.

\subsection{Notations} 
A summary of the notations is listed in Tab.~\ref{tab:notations}.

\begin{table}
    \caption{Choices of hyper-parameters for training.}
    \footnotesize
    \centering
    \begin{tabular}{lc}
	\toprule
 Name  & Value \\
 \midrule
  \multicolumn{2}{c}{\bf Optimization parameters}\\
 Network architecture of $\phi_w$ & ResNet-18 (ImageNet-pretrained)~\cite{he2016deep}\\
 Optimizer          &   Adam~\cite{kingma2014adam}\\
 Learning rate for $\phi_w$             &   $1\times 10^{-4}$\\
 Learning rate for other params.     &   $5\times 10^{-3}$\\
 Batch size         &   8 image pairs\\
 Loss weight $\{\beta_1,\dots,\beta_4\}$    & $\{0.5, 0.5, 2, 5\times 10^{-3}\}$\\
 \midrule
 \multicolumn{2}{c}{\bf Measurement pre-processing}\\
 Crop center        & Center of object bounding box\\
 Crop size          & 1.2 $\times$ longest edge\\
 Resized to          & $256 \times 256$\\
 \bottomrule
\label{tab:hp}
\end{tabular}
\vspace{-10pt}
\end{table}

\begin{table*}
    \caption{Table of notations used in this work.}
    \centering
    \begin{tabular}{lll}
	\toprule
 Symbol  & Description \\
 \midrule
 \multicolumn{2}{c}{\bf Numbers}\\
 $T$        &   Number of frames in the input video\\
 $M$        &   Number of faces in the mesh\\
 $N$        &   Number of vertices in the mesh\\
 $B$        &   Number of bones for LBS\\
 \midrule
 \multicolumn{2}{c}{\bf Measurements}\\
 $I_t$      &   Input RGB image at time $t$\\
 $S_t$      &   Input or measured object silhouette image at time $t$\\
 ${\bf u}^{+}_t$    &   Input or measured forward optical flow map from time $t$ to $t+1$\\
 ${\bf u}^{-}_t$    &   Input or measured backward optical flow map from time $t$ to $t-1$\\
 ${\bf Y^*}$        &   Union of all measurements \{$I_t$, $S_t$, ${\bf u}^{+}_t$, ${\bf u}^{-}_t$\}\\
 \midrule
 \multicolumn{2}{c}{\bf Renderings}\\
 $\hat{I}_t$      &   Rendered color image of the object at time $t$\\
 $\hat{S}_t$      &   Rendered object silhouette image at time $t$\\
 ${\bf \hat{u}}^{+}_t$    &   Rendered forward optical flow map of the object from time $t$ to $t+1$\\
 ${\bf \hat{u}}^{-}_t$    &   Rendered backward optical flow map of the object from time $t$ to $t-1$\\
 ${\bf Y}$        &   Union of all renderings \{$\hat{I}_t$, $\hat{S}_t$, ${\bf \hat{u}}^{+}_t$, ${\bf \hat{u}}^{-}_t$\}\\
 \midrule
 \multicolumn{2}{c}{\bf Variables}\\
 $f_t$                  & Focal length of the camera at time $t$\\
 ${\bf K_t}$            & Intrinsic matrix of a simple pinhole camera (with zero skew and square pixel) at time $t$ \\
 ${\bf R_{0,t}}$        & Object root body rotation matrix $\in SO(3)$ at time $t$ \\
 ${\bf T_{0,t}}$        & Object root body translation vector at time $t$ \\
 ${\bf G_{0,t}}$        & Object root body transformation at time $t$, ${\bf G_{0,t}}=\begin{pmatrix}[c|c]{\bf R_{0}} & {\bf T_{0}}\end{pmatrix}{\bf_t}$ \\
 ${\bf R_{1\dots B, t}}$   & Bone rotations from the rest pose to time $t$\\
 ${\bf T_{1\dots B, t}}$   & Bone rotations from the rest pose to time $t$\\
 ${\bf G_{1\dots B, t}}$   & Bone transformations from the rest pose to time $t$, ${\bf G_{i,t}}=\begin{pmatrix}[c|c]{\bf R_{i}} & {\bf T_{i}}\end{pmatrix}{\bf_t}, i\in \{1\dots,B\}$ \\
 ${\bf D_t}$                & Union of camera and bone transformations $\{\bf G_{0,t}, \dots, G_{B, t}\}$\\
 ${\bf P_t}$            & Projection matrix of the camera at time $t$, ${\bf P_t}={\bf K_t}{\bf G_{0,t}}$\\
 $\Delta{\bf V_t}$      & Vertex motion from the rest shape to time $t$  \\
 \midrule
 \multicolumn{2}{c}{\bf Parameters}\\
 $(p_x,p_y)$                    & Principal point of the camera \\
 ${\bf \bar{V}}_i$                & Position of the i-th vertex of the mesh in the rest pose (or mean shape)\\
 ${\bf \bar{C}}_i$                & Color of the i-th vertex of the mesh \\ 
 ${\bf S}$                      & Union of all mesh parameters, ${\bf S}=\{{\bf \bar{V}}, {\bf \bar{C}}, {\bf F}\}$\\
 ${\bf J}_b$                      & Position of the center of the b-th bone (or Gaussian component) \\
 ${\bf Q}_b$                      & Precision matrix of b-th bone (or Gaussian component)\\
 ${\bf W}$                      & Skinning weights matrix, {\bf W}=$\{ {\bf J}, {\bf Q} \}$\\
 $\phi_w$                       & Weights of the convolutional camera and pose network\\
 ${\bf n^*}$                    & Normal vector of the symmetry plane in the canonical frame\\
 ${\bf X}$              & Union of all Parameters  \\
 \midrule
 \multicolumn{2}{c}{\bf Constants}\\
 ${\bf F}$                      & Faces of the mesh\\
 ${\boldsymbol \beta}$          & Weights between the losses\\
 ${\bf H}$                      & Householder transformation matrix describing reflection about the y-z plane\\
 ${\bf n}_0$                    & Unit vector towards to the x axis\\
\midrule
 \multicolumn{2}{c}{\bf Others}\\
 {\bf S0}       &  Training stage 0: optimize for $\{\phi_w({f_t, \bf G_{0,t}}), p_x, p_y, {\bf n}^*, {\bf \bar{V}}, {\bf \bar{C}}\}$\\
 {\bf S1-3}     &  Training stage 1 to 3: optimize for $\{\phi_w(f_t, {\bf G_{ {\bf 0 \dots B},t}}), p_x, p_y, {\bf n}^*, {\bf \bar{V}}, {\bf \bar{C}}, {\bf J}, {\bf Q}\}$\\
 \bottomrule
\label{tab:notations}
\end{tabular}
\vspace{-10pt}
\end{table*}

\end{document}